%% file: acl2021.tex
\documentclass[11pt,a4paper]{article}
\usepackage[hyperref]{acl2021}
\usepackage{times}
\usepackage{url}
\usepackage{booktabs,siunitx}
\usepackage{amsfonts}
\usepackage{amsmath}
\usepackage{subfigure}
\usepackage{graphicx}
\usepackage{latexsym} 
\usepackage{multirow}

\usepackage{microtype}

\aclfinalcopy 
\def\aclpaperid{502} 

\title{A Dialogue-based Information Extraction System for Medical \\Insurance Assessment}

\author{Shuang Peng\thanks{~~Equal contributions.}  \quad Mengdi Zhou$^{*}$ \quad Minghui Yang \quad Haitao Mi \\ { \bf Shaosheng Cao \quad Zujie Wen \quad Teng Xu  \quad Hongbin Wang \quad Lei Liu} \\
Ant Group \\
Hangzhou, China\\
{\tt \{jianfeng.ps,jacquelyn.zmd,minghui.ymh,haitao.mi,shaosheng.css,} \\
{\tt zujie.wzj,harvey.xt,hongbin.whb,leihu.ll\}@antgroup.com}}

\date{}

\begin{document}
\maketitle
\begin{abstract}
In the Chinese medical insurance industry, the assessor's role is essential and requires significant efforts to converse with the claimant.
This is a highly professional job that involves many parts, such as identifying personal information, collecting related evidence, and making a final insurance report. 
Due to the coronavirus (COVID-19) pandemic, the previous offline insurance assessment has to be conducted online. 
However, for the junior assessor often lacking practical experience, it is not easy to quickly handle such a complex online procedure, yet this is important as the insurance company needs to decide how much compensation the claimant should receive based on the assessor's feedback. 
In order to promote assessors' work efficiency and speed up the overall procedure, in this paper, we propose a dialogue-based information extraction system that integrates advanced NLP technologies for medical insurance assessment.
With the assistance of our system, the average time cost of the procedure is reduced from 55 minutes to 35 minutes, and the total human resources cost is saved 30\% compared with the previous offline procedure. Until now, the system has already served thousands of online claim cases.
\end{abstract}

\section{Introduction}
\input{intro.tex}

\section{System Framework and Technique Details}
\input{system.tex}

\section{Experiments}
\input{experiment.tex}

\section{Deployment Details}
\input{eval.tex}

\section{Discussions on Technique Limitations}
\input{discussion.tex}

\section{Related Work}
\input{related.tex}

\section{Conclusions and Future Work}
In this paper, we presented a dialogue-based information extraction system that helps the insurance assessor better perform their job. The system combines different NLP technologies like sentence similarity learning, NER/EL, and DST methods in a novel way. While there has been considerable work done for each of the independent modules/models, 
this systematic way of combining them enables the platform to quickly deliver what may be of significant value to the insurance industry. 

In the future, we would like to try to replace the current pipeline method with the end2end method based on text-generation when the accumulated annotated data is enough.

\section{Ethical Considerations}
\input{ethical.tex}

\section*{Acknowledgements}
The authors would like to thank Gan Li, Jianzeng Liang, Chaojin Zhou, Weiwei Fan, Li Qiao, and other members of Ant Group for helpful discussions and comments. We would also like to thank reviewers for their valuable comments.

\bibliography{anthology,acl2021}
\bibliographystyle{acl_natbib}

\end{document}

%% file: intro.tex
In the Chinese medical insurance industry, the assessor's role is essential in handling protection claims. 
The insurance assessor may work for the insurance company or be a third-party administrator assessor.
The claimant approaches the insurance assessor to evaluate the disease that has affected them. 
For example, if the claimant has lung cancer, the insurance assessor would check the medical cost and decide if they are claimable. 
Once the assessor has finished collecting the evidence and related claimant information 
(i.e., disease history, medication experience, in-hospitalized information, etc.), a final detailed case report will be made and sent to the insurance company.

The traditional insurance assessment procedure is carried out offline where the assessor conducts an onsite interview with the claimant. 
Due to the coronavirus (COVID-19) pandemic, the previous offline insurance assessment has to be interrupted temporarily, 
and thus many claimants can not get their insurance compensation in time. Some of them, meanwhile, may also suffer job loss\footnote{According to the statistics result from the United Nations, COVID-19 could cause the equivalent of 400 million job losses in the second quarter of 2020.}. To help these claimants, many insurance companies shift the offline procedure to the online platform in which insurance assessors can communicate with claimants via video connection. 
In this way, claimants can submit their applications and get compensations entirely online. On the other hand, the online assessment also helps the insurance company save expenses as the assessor can quickly approve qualified applications or screen out applications that do not conform to the claims on the online platform.

\begin{figure*}[ht]
\centering
\includegraphics[width=1.35\columnwidth]{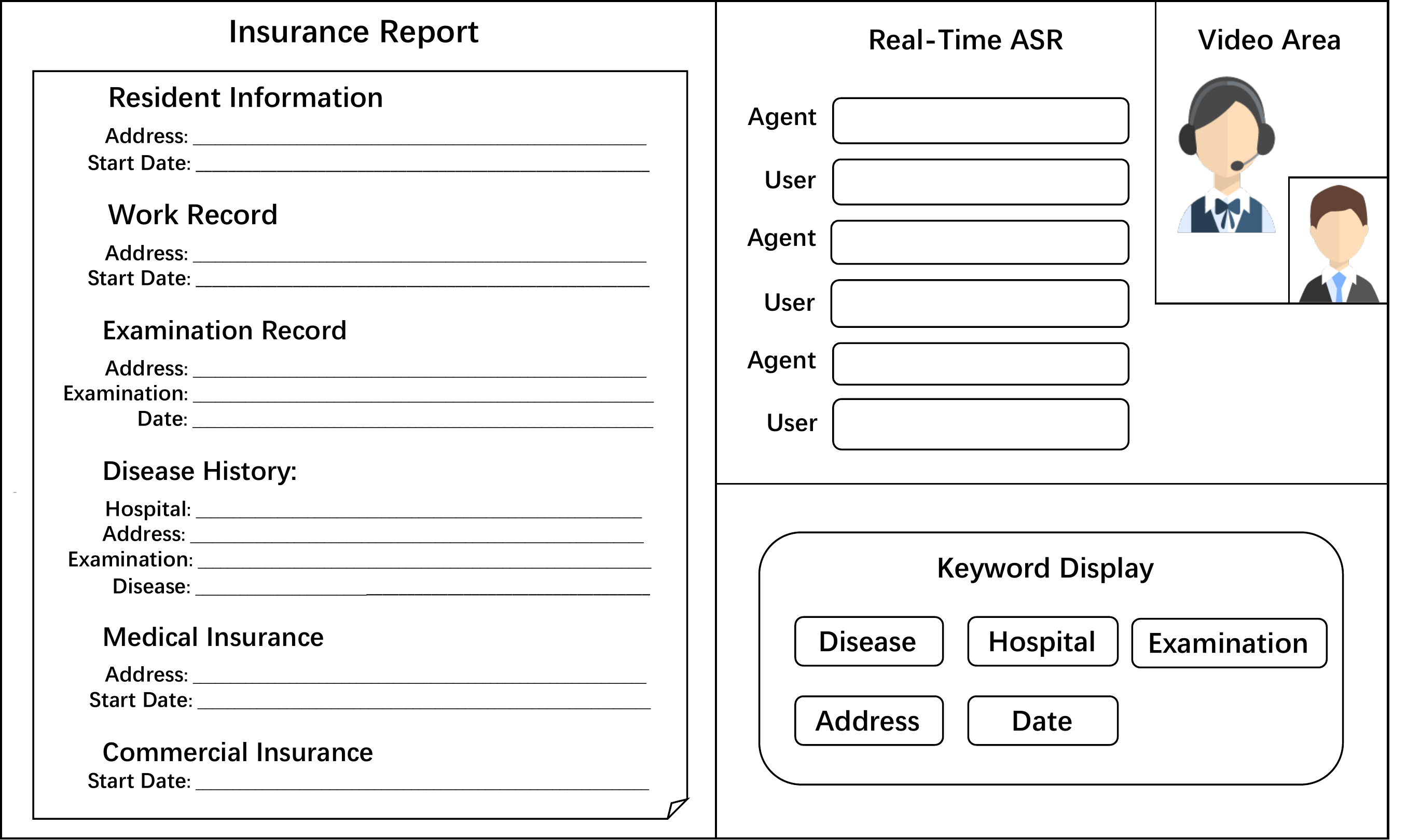}
\caption{The illustration of the front end of Intelligent Insurance Assessment System.}
\label{fig:framework}
\end{figure*}

However, limited by the complexity of the insurance assessment procedure and the instability of the online environment, simply providing a video connection for insurance assessors and claimants is still insufficient. We summarize three primary challenges for the current online procedure: (1) Instability. The online video connection may sometimes be unstable, and the assessor may not distinguish the claimant's words. (2) Complexity. The insurance assessment procedure is very complex, and not all assessors are sufficiently experienced. The junior assessors, in particular, need much assistance to complete their work expertly, and even some senior assessors still have room for improvement. (3) Time-consuming. Recording the key information and writing the insurance report is tedious for most assessors, and junior assessors may even forget some important information during the inquiry.

To this end, we propose an \textbf{I}ntelligent \textbf{I}nsurance \textbf{A}ssessment \textbf{S}ystem, called \textbf{IIAS} (shown in Figure~\ref{fig:framework}) for medical insurance assessment.
The system aims at promoting the work efficiency of the insurance assessor through dialogue-based information extraction. For these purposes, We use recent NLP technologies, such as streaming automatic speech recognition (ASR)~\cite{moritz2020streaming, mani2020asr}, large scale pre-trained language models~\cite{devlin2018bert, chinesebertwwm}, sentence similarity learning~\cite{chen2017enhanced, enhancedrcnn}, named entity recognition/linking (NER/EL)~\cite{le2018improving,devlin2018bert} and dialogue state tracking (DST)~\cite{ouyang2020dialogue,zhang2020recent}, to ensure high performance while requiring only a small amount of annotated data. 

Our IIAS alleviates the cognitive workload of assessors in several steps: 
(1) Our streaming ASR transforms speech signal into the conversation text in real-time, and our sentence similarity learning method labels corresponding topics for each conversation on-the-fly. 
(2) Our NER component then extracts raw entities from the real-time conversation text, and the EL part links the raw entities into the unified insurance knowledge base (KB) for getting all possible keywords. 
(3) Our DST method, including question and negation identification modules, tracks the state of dialogue context for filtering irrelevant keywords.

To the best of our knowledge, this is the first work to propose an intelligent system for insurance assessment. The main contributions of our work are concluded as follows.
\begin{itemize}
\item We propose a dialogue-based information extraction system that shifts the previous insurance assessment procedure online and provides necessary intelligent assistance for the assessor.
\item We apply recently advanced NLP technologies to address the problem of online insurance assessment. Our methods significantly improve the assessor's work efficiency, where the average time cost of the insurance assessment procedure is reduced from 55 minutes to 35 minutes.
\item Our system has already been deployed in the real world. Until now, it has served thousands of online insurance claim cases, which delivers potential value to the insurance industry. 
\end{itemize}

%% file: system.tex
\begin{figure*}[t]
\centering
\includegraphics[width=2\columnwidth]{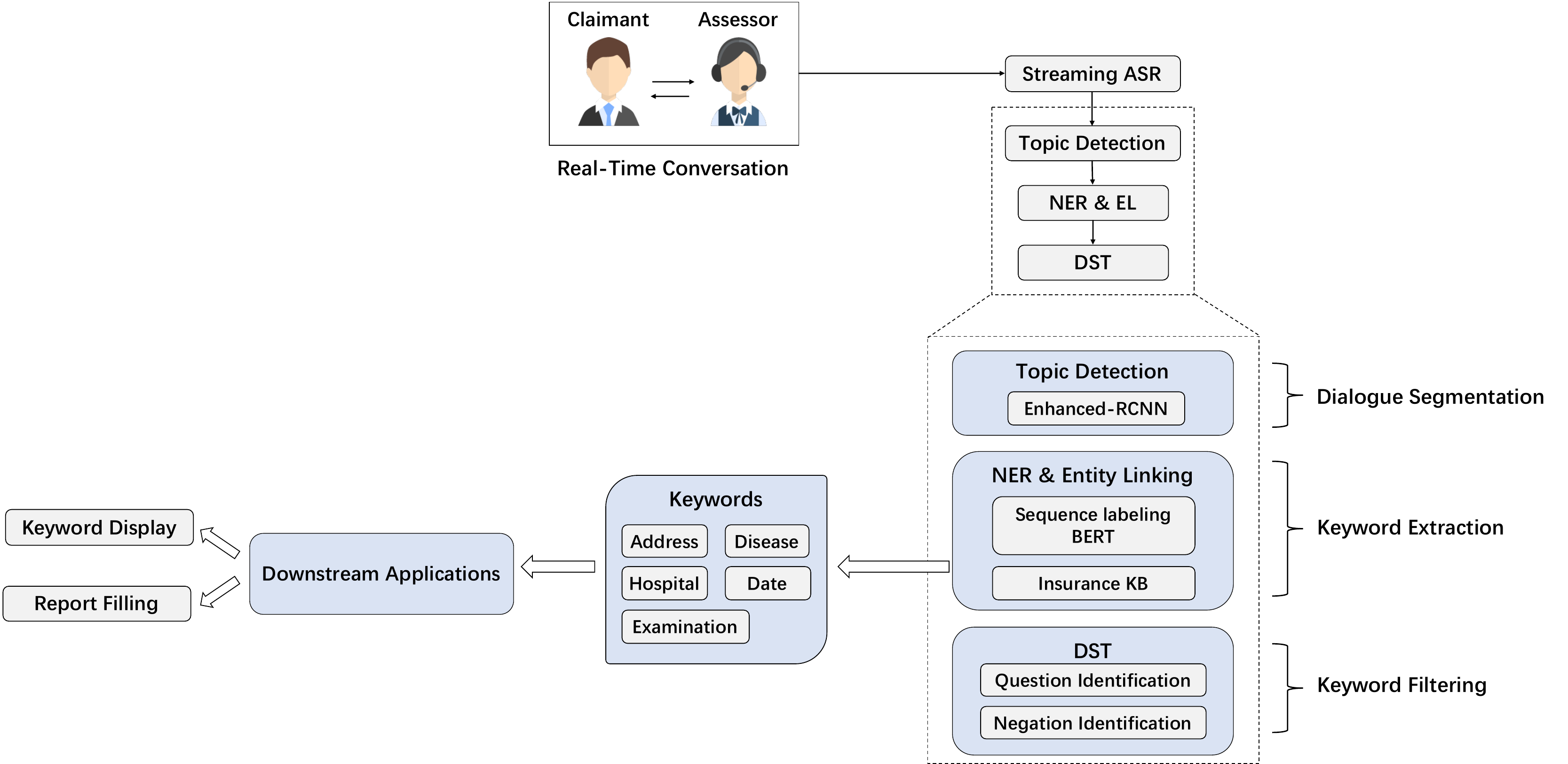}
\caption{The Architecture of Intelligent Insurance Assessment System.}
\label{fig:flow}
\end{figure*}

Figure~\ref{fig:flow} shows the architecture of IIAS with key components such as streaming ASR, dialogue segmentation, keyword extraction (NER and EL), keyword filtering (question and negation identification), and downstream applications (keyword display and report filling).

As previously mentioned, recording the information during the assessment is tedious for most insurance assessors. In order to increase the efficiency of information collection and reduce the assessor's cognitive workload, IIAS provides a keyword-based feature that first displays keywords on the system dashboard and then intelligently suggests the related content when assessors are filling in the insurance report. 

For better understanding, we discuss the technical details for each key component in the following subsections. Besides, we also present the partial experimental result corresponded to each component.

\subsection{Dialogue Segmentation}
Dialogue segmentation aims at detecting the topic of the real-time conversation text. As the insurance assessment procedure is conducted according to the content that needs to be filled in the insurance report, different dialogue between the assessor and claimant is focused on different topics. These topics can be seen as different questions that the claimant needs to answer. For example, the assessor may start with the topic about the claimant's work address, and then inquire about when the disease starts and where the claimant receives the treatment, and finally inquire about the doctor's advice and in-hospitalized information. Here we use the sentence similarity learning method to map the assessor utterance into the corresponding topic.

First, we order the one-to-many mapping relationship between the topic and standard questions that most assessors inquire in the insurance assessment procedure. We select the representative utterance from the assessor as the standard question. 
The standard question is a set of questions that represent different topics across all assessors. We select the standard question by running HDBSCAN clustering algorithm~\cite{mcinnes2017hdbscan} followed with a human adjustment. 
Since the ways of inquiry among assessors are different, and the ASR results are sometimes affected by the outside speech signals, we use the Enhanced-RCNN model~\cite{enhancedrcnn} to calculate the similarity score between the utterance and selected standard questions, and get the topic of the utterance if the score is greater than the threshold (the default value is 0.5) or follow the topic of the previous utterance if the score is less than the threshold.

After the dialogue segmentation, the whole dialogue is segmented into many parts that belong to different topics so that the keywords extracted in the following process can be linked with these topics in real time. We evaluate the performance of dialogue segmentation by 200 online cases with the human annotation. The accuracy of segmentation reaches 90\%.

\subsection{Keyword Extraction}
In the insurance assessment scenario, the keyword contains five types\footnote{Address, Hospital, Disease, Date, and Examination}, and their values are not within a limit number. Therefore, any string fragment of assessor or claimant utterance may become a keyword. Traditional slot filling methods with classification models in the task-oriented dialogue system cannot predict unseen slot values in a pre-defined value list and are not appropriate to our scenario. To solve the problem, we convert the keyword 
\begin{figure}[ht]
 \centering
 \includegraphics[width=1\columnwidth]{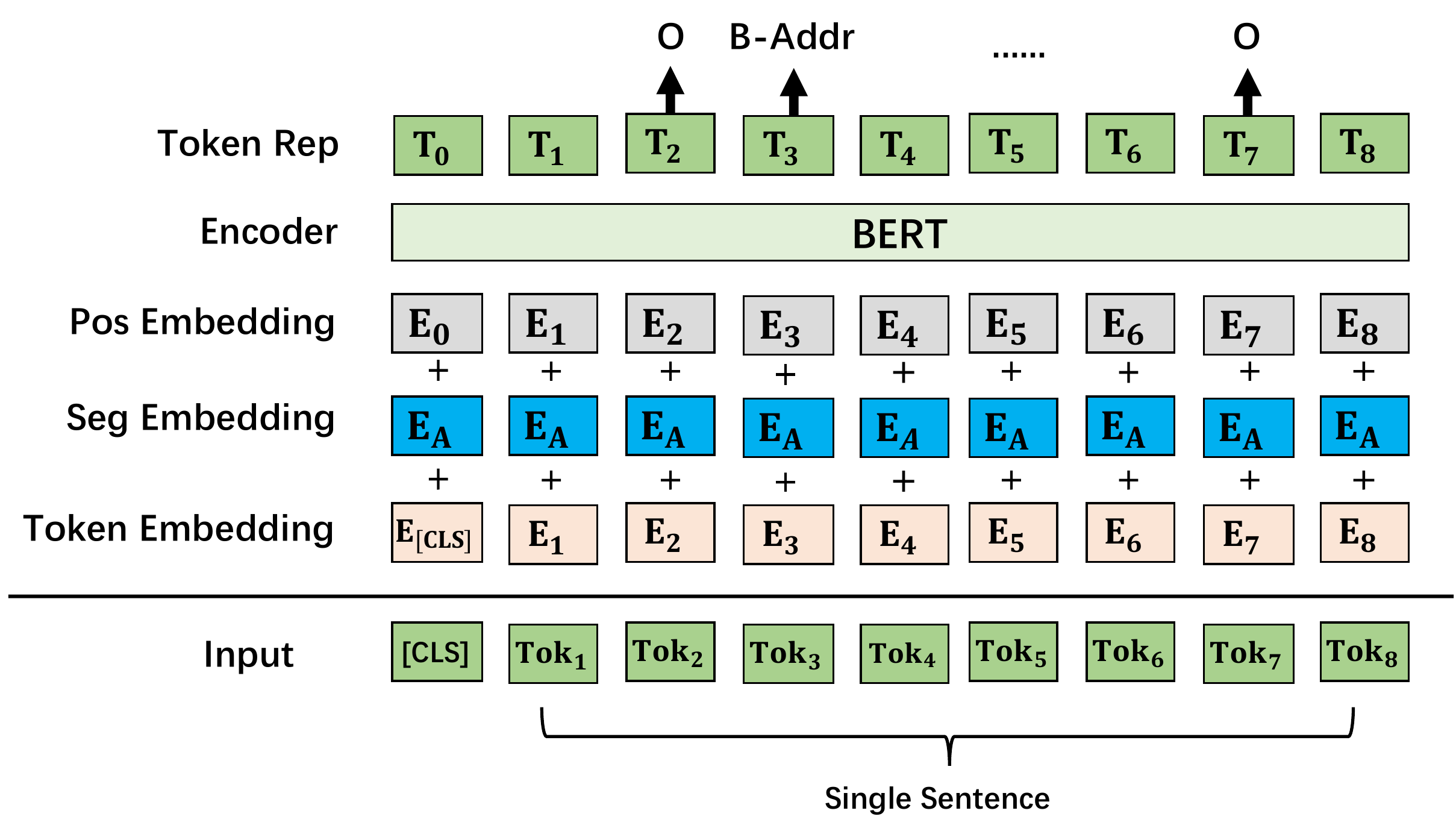}
 \caption{The sequence labeling BERT model for NER.}
 \label{ner_model}
\end{figure}
extraction as a sequence labeling task and use the BERT-based NER model~\cite{devlin2018bert}.

We first pre-annotate the keyword type in the utterance as \texttt{Addr}, \texttt{Hos}, \texttt{Dis}, \texttt{Date}, \texttt{Exam} or \texttt{Other} (non-named entity) and follow the official tutorial in NER\footnote{https://github.com/google-research/bert}. The visual representation of BERT-based NER model is shown in Figure~\ref{ner_model}. For fine-tuning, we feed the final hidden representation for each token into a classification layer over the NER label set, and the predictions are non-autoregressive and no CRF~\cite{devlin2018bert}. After the NER process, we get the raw entities extracted from the utterances. As the word accuracy of ASR is about 85\%, some of the derived entities may be misrecognized (correct entity type but with a wrong form).
In order to alleviate the negative effect of the ASR module, we use the EL model~\cite{le2018improving} to recognize and disambiguate the raw entities to the insurance KB that contains the normalized address, hospital and disease information for getting all possible keywords. 

\begin{figure*}[ht]
 \centering
 \includegraphics[width=2\columnwidth]{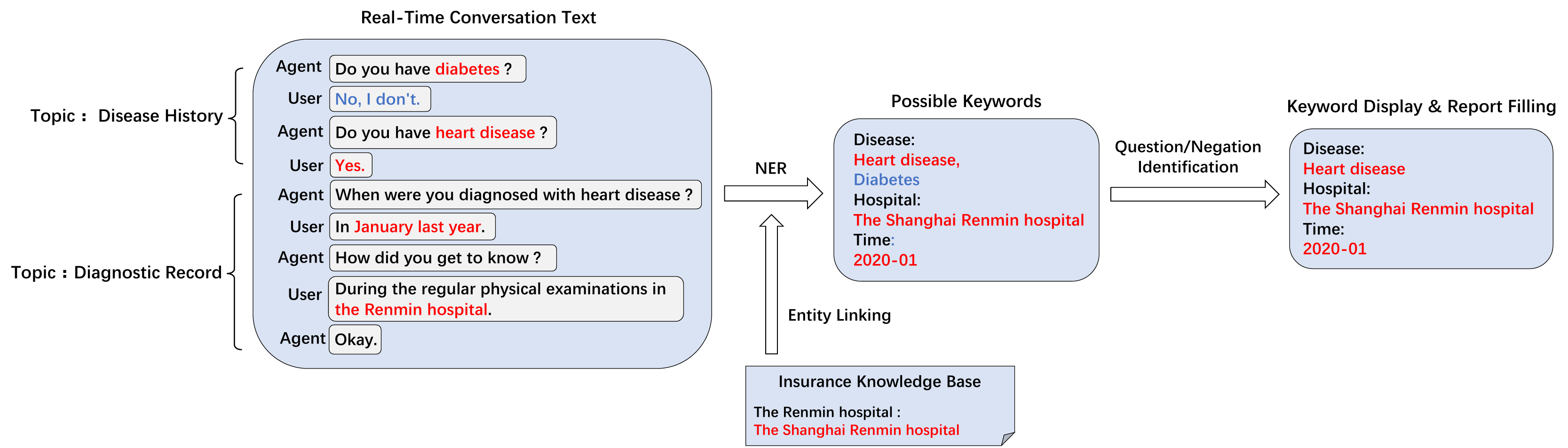}
 \caption{An example of IIAS workflow. In this example, the real-time conversation text is segmented into two topics, and keywords extracted from the utterances are first filtered, and then displayed on the system dashboard, and finally suggested when the assessor fills in the insurance report.}
 \label{workflow}
\end{figure*}

\begin{table}[]
\setlength{\tabcolsep}{1mm}
\centering
\begin{tabular}{l|ccccc}
\toprule
\textbf{Entity Type}&\textbf{Addr.}&\textbf{Hos.}&\textbf{Dis.}&\textbf{Date.}&\textbf{Exam.}\\
\hline
F1-Score (\%)&78.2&78.3&83.1&85.1&85.4\\
\bottomrule
\end{tabular}
\caption{Test performance of the NER/EL model on five entity types.}
\label{table:ner_exp}
\end{table}

To measure the NER/EL model's performance separately, we take the F1-Score as the evaluation metric. The experimental results on the human-annotated testing set are shown in Table~\ref{table:ner_exp}. The results show that the NER/EL models achieve good performance on all five entity types as the average F1-score has reached 82\%. 

\subsection{Keyword Filtering}
During the insurance assessment procedure, the assessor always asks the claimant many detailed questions.
As a result, a part of entities extracted by the previous keyword extraction process may come from the assessor's question text and actually do not need to be displayed or suggested if the claimant's following response has negative semantics.
To filter the irrelevant keywords extracted by the previous process and increase the accuracy of keyword display and report filling, we design a DST method for keyword filtration. Like the application in the task-oriented dialogue system, the DST method tracks the keyword's state (addition or deletion) and contains two major modules: question identification and negation identification. 
The assessor's conversion text is first fed to the question identification module.
Suppose the text is identified to be a question. 
In that case, the extracted keyword from it will be reserved temporarily and only displayed (or recommended) once they are confirmed by the claimant (decided by the negation identification module).


\subsubsection{Question Identification}
Question identification determines whether the current utterance belongs to part of the insurance assessor's inquiry. We convert question identification as the binary classification task and utilize the text classification model to solve it.

We utilize the LSTM~\cite{hochreiter1997long}, one of the most popular text classification models, as the base model. In this case, the input 
\begin{math}
  x = \{x_1, x_2, ..., x_T\}
\end{math}
is the text transformed by the ASR module where $x_t$ is the word after segmentation. The output is a label $y$ indicating whether the text is a question or not. 

Based on the observation, the insurance assessment's topic distribution is diverse, and the accuracy of question identification is highly relevant to the concrete topic (i.e., resident/work address,  medication experience, or disease history). Therefore, training a single model for covering all the topics is actually not the best solution in our scenario, and it is necessary to train different identification models for different topics. However, the total amount of manually labeled data is not large enough to train multiple classifiers for every topic. To alleviate this issue, we adopt the adversarial multi-task training method \cite{liu2017adversarial, yang2020servicegroup}. Compared with the multi-task training method, the adversarial-based training method improves the question identification model's performance for general purposes and a particular topic with only a few labeled data.

\begin{table}
  \centering
  \begin{tabular}{l|c}
    \toprule
  \textbf{Model}&\textbf{Accuracy (\%)}\\
    \hline
    Single Model&61.2\\
    Multi-task Model&75.1\\
    Adversarial Multi-task Model&\textbf{78.4}\\
  \bottomrule
\end{tabular}
\caption{Test Performance on Question Identification.}
\label{tab:adv}
\end{table}

Experimental results on different training methods are presented in Table~\ref{tab:adv}.
The results show that the adversarial multi-task model achieves the best accuracy among three compared models, which indicates that the adversarial learning method is more appropriate to the insurance scenario that the topic distribution of question is unbalanced.

\subsubsection{Negation Identification}
After the question identification module, the following claimant's utterance is feed to the negation identification module. 
The keywords extracted from the question are only displayed on the system dashboard if the claimant confirms.
Here we use the BERT-based binary-classification model~\cite{devlin2018bert} for negation identification. The input is the current claimant utterance with the previous two rounds of utterances from the assessor and claimant (we use \texttt{[SEP]} to separate them).  The output is the label indicating whether the current utterance has negative semantics or not (the final layer is a two-class softmax layer).
From the performance on the human-annotated testing set, the negation identification module's accuracy is 92\%.

The ablation study on the DST method's contribution (question and negation identification modules) is present in the experiment section.

\subsection{Applications}
Based on the previous three steps, the system gets filtered keywords extracted from the real-time conversion text, and these keywords are classified into different topics simultaneously. 

To help the assessor better do their job, we identify two types of generic applications for derived keywords: keyword display and report filling. The keyword display aims at displaying information extracted from the real-time conversation text on the system dashboard. Furthermore, the report filling aims at recommending related contents when the assessor fills in the specific part of the insurance report. For ensuring a good user experience, we sort extracted keywords by the reversed order of utterances in the dialogue and limit the maximal number of recommended contents as 5. We show an example to illustrate the IIAS workflow in Figure~\ref{workflow}. 
The figure shows how IIAS extracts the information from the real-time conversation text and uses them for the downstream applications.

For the insurance assessor, keyword display and report filling significantly alleviate the workload of remembering all the claimant information during the inquiry procedure. To further evaluate IIAS's overall contribution to the assessor's work, we conduct different experiments in the experiment section.

%% file: experiment.tex
\begin{table}[]
    \centering
    \begin{tabular}{l|rr}
    \toprule
    \textbf{Items}               & \textbf{Train}          & \textbf{Test}  \\ \hline
    Num of dialogs               & 5000                    & 200            \\ 
    Average Rounds Per Dialog    & 497.6                   & 501.2          \\ 
    Average Keywords Per Dialog  & 25.3                    & 27.9          \\ 
    \bottomrule
    \end{tabular}
    \caption{Statistics of Insurance Assessment Dataset.}
    \label{table:insurance}
\end{table}

\begin{table*}[]
\centering
\scalebox{1}{
\begin{tabular}{l|c|c|c|c|c|c|c}
\toprule
\multicolumn{1}{l|}{\textbf{Topic}} & \multicolumn{2}{c|}{\textbf{Resident Information}} & \multicolumn{2}{c|}{\textbf{Work Record}} & \multicolumn{3}{c}{\textbf{Diagnostic Record}} \\ \hline
\textbf{Entity Type} & \multicolumn{1}{c|}{\textbf{Addr.}} & \multicolumn{1}{c|}{\textbf{Date.}} & \multicolumn{1}{c|}{\textbf{Addr.}} & \multicolumn{1}{c|}{\textbf{Date.}} & \multicolumn{1}{c|}{\textbf{Date.}} & \multicolumn{1}{c|}{\textbf{Hos.}} & \multicolumn{1}{c}{\textbf{Exam.}} \\ \hline
Retrieval System   & 85.6 & -   & 84.2 & -   &-   & 67.9 & 65.7 \\ \hline
IIAS & 83.5 & 43.3 & \textbf{85.7} & 55.4 & 52.1 & \textbf{74.0} & \textbf{68.8} \\ 
\bottomrule
\end{tabular}%
}
\caption{The Recall@5(\%) of report filling on different topics (resident information, work record and diagnostic record).}
\label{table:report_filling_exp1}
\end{table*}

\begin{table*}[]
\centering
\scalebox{1}{
\begin{tabular}{l|c|c|c|c|c|c|c}
\toprule
\multicolumn{1}{l|}{\textbf{Topic}} & \multicolumn{4}{c|}{\textbf{Disease History}} & \multicolumn{2}{c|}{\textbf{Medical Insurance}} & \multicolumn{1}{c}{\textbf{Commercial Insurance}} \\ \hline
\textbf{Entity Type} & \multicolumn{1}{c|}{\textbf{Date.}} & \multicolumn{1}{c|}{\textbf{Hos.}} & \multicolumn{1}{c|}{\textbf{Exam.}} & \multicolumn{1}{c|}{\textbf{Dis.}} & \multicolumn{1}{c|}{\textbf{Addr.}} & \multicolumn{1}{c|}{\textbf{Date.}} & \multicolumn{1}{c}{\textbf{Date.}} \\ \hline
Retrieval System    & - & 68.3 & 71.5 & 25.7 & 92.3 & -  & -  \\ \hline
IIAS  & 57.9 & \textbf{74.8} & \textbf{76.9} & \textbf{28.8} & 91.8 & 54.9 & 46.7 \\ \bottomrule
\end{tabular}
}
\caption{The Recall@5(\%) of report filling on different topics (disease history, medical insurance and commercial insurance).}
\label{table:report_filling_exp2}
\end{table*}

\subsection{Insurance Assessment Dataset}
We manually construct an insurance assessment dataset from the real-world environment and use it for evaluating the performance of IIAS. The dataset consists of 200 online cases, and each case contains two parts: a multi-turn dialogue and a corresponding insurance report that the assessor needs to finish. To ensure the high quality of the dataset, we employ five real insurance assessors to fill in the insurance report by selecting a string fragment of the utterance from the given multi-turn dialogue. Here we screen out forms in the insurance report that cannot be filled based on the given multi-turn dialogue. Table~\ref{table:insurance} shows the detailed statistics of the insurance assessment dataset.

In order to protect data privacy, the usage of the claimant's private data must be authorized. We have obtained explicit permissions from the claimant to use their personal private data before the insurance assessment. If a claimant disapproves of the authorization, we will not use her/his data. For the authorized user privacy data, we have a series of strict processes to ensure the data remain confidential.

\subsection{Experimental Setup}
\subsubsection{Method in Comparison}
As far as we know, there is no similar research work in the insurance industry. To examine the effectiveness of IIAS, we compare it with the retrieval-based system on the report filling performance. Different from IIAS that is mainly based on the real-time ASR and speaker recognition, the retrieval system suggests retrieved results from the pre-constructed KB based on the conversation text. For example, the system will recommend \textsl{top-k} relevant disease nouns in the KB when the assessors fill in the form about what disease the claimant is diagnosed with. To be aligned with IIAS, we limit the retrieved number \textsl{k} in the retrieval system as 5. 

The major limitations of the retrieval system are that the content that needs to be filled may only exist in the conversation text but not appear in the pre-constructed KB. 

\subsubsection{Evaluation Metrics}
The evaluation process is divided into offline and online parts.

\textbf{Offline Evaluation:} We choose the recall as the metric to evaluate the system performance of report filling. The recall measures the ratio of suggested correct content to all ground-truth content that needs to be filled.

\textbf{Online Evaluation:} We choose the time-saving and efficiency-improving as metrics because our work's primary purpose is to alleviate assessors' workload and reduce the insurance assessment procedure's time cost.

\subsection{Evaluation Results and Discussions} 
The experimental results\footnote{To make a fair comparison, we do not compare the recall ratio on the \texttt{Date} type because the value of \texttt{Date} cannot be included by the pre-constructed KB in the retrieval system.} on the report filling performance between IIAS and the retrieval system are shown in Table~\ref{table:report_filling_exp1} and Table~\ref{table:report_filling_exp2}.

We summarize our observations as follows: 

(1) Our IIAS is generally better than the retrieval system on the record filling's performance for most topics. This shows that using the method based on speaker recognition is more effective with less dependence on the pre-constructed KB.

(2) For entities of \texttt{Addr} type, the retrieval system achieves comparative performance than IIAS. This indicates that the address entities are relatively fixed and easy to be included in the pre-constructed KB. For this type of entity, the retrieval system is more appropriate.

(3) For entities of \texttt{Exam} and \texttt{Dis} types, our system achieves much better performance than the retrieval system. This shows that for those entity types that are diverse and hard to be included by the pre-constructed KB, using the NER/EL methods to extract keywords from the utterances is more appropriate.


\subsection{Ablation Study}
\begin{table}[]
\setlength{\tabcolsep}{1.6mm}
\centering
\begin{tabular}{l|c|c}
\toprule
\textbf{Entity Type}&\textbf{Hospital}&\textbf{Disease}\\
\hline
w/o DST&71.79&39.20\\
\hline
Complete System&\textbf{77.31}&\textbf{53.24}\\
\bottomrule
\end{tabular}
\caption{Evaluation results of system ablation on the Precision(\%) of keyword extraction for \texttt{Dis} and \texttt{Hos} types.}
\label{table:keyword_recommend}
\end{table}

We perform the ablation study to validate the contribution of the DST method (question and negation identification). Since the DST method filters out irrelevant keywords extracted by the NER/EL modules, we choose the \textsl{hospital} and \textsl{disease} types that frequently appear in the assessor's utterances for the experiment. 

The experimental results are presented in Table~\ref{table:keyword_recommend}. 
Here we evaluate the precision of the keyword extraction.
The results show that with the addition of the DST method, the keyword extraction's precision is significantly promoted.
This indicates that the DST method effectively filters the error keywords, which is beneficial for improving system performance.

\subsection{Online Performance}
From the statistical result for three months after the system deployment, the average time cost of the insurance assessment procedure is reduced from 55 minutes to 35 minutes, and the overall human resources cost is saved 30\% compared with previous offline insurance assessment.

Moreover, with the assistance of our IIAS, the insurance company provides more job opportunities for junior assessors. This indicates that IIAS has effectively lowered the bar of insurance assessment jobs.

%% file: eval.tex
Our proposed approach consists of different NLP components that are trained individually. We deploy them on two GPUs (Tesla P100 16G) and provide a unified HTTP interface for the back-end system.

To support multiple users, the computation cost needs to be well distributed across different GPUs.
For each GPU, we set up two independent processes. A load-balancing module is then employed to distribute the users' requests to different processes based on the working loads. 
Based on the stress test, the online service's average response time is less than 150ms and can support a peak QPS of 35.

%% file: discussion.tex
In this section, we discuss the limitations of techniques in IIAS.
Based on the observation of system results and feedback from the insurance assessors, the limitations are from three parts: 

(1) \textbf{ASR Results}: The performance of two applications (keywords display and report filling) heavily relies on the ASR module, but the ASR errors are the major problem of our system. The ASR error may result in correct entity type but with a wrong form, which indicates that optimizing the ASR module should be a top priority for future research.

(2) \textbf{Reasoning Problem}: The current system cannot solve the problem that requires complex reasoning, such as date recognition (e.g., last Sunday or three days ago.). However, many date recognitions during the insurance assessment procedure require reasoning. As a result, we plan to include reasoning ability into the system in the future.

(3) \textbf{Accumulative Errors}: The current system is based on a multi-model approach where different NLP components such as topic detection, NER/EL, and question/negation identifications are trained individually and combined.
This type of pipeline approach is easy to lead to accumulative errors. We have tried to replace the current approach with the end-to-end summarization method based on text-generation. 
However, text-generation models always require many training data and their predictions are sometimes uncontrollable (the inference time is also slow if we use the pre-trained model like GPT-2~\cite{radford2019language}). 
Although now the end-to-end method cannot be applied in the real-world setting (especially in our insurance scenario that requires controllability), it has some advantages and is still part of our plan. 

For these limitations, we are working on a new solution to improve the current system.

%% file: related.tex
With the advances in NLP, agent-assist systems have been used in various domains, including technical support, reservation systems, and banking applications ~\cite{fadnis2020agent}. In this section, we briefly introduce some related technologies used in IIAS.

\subsection{Sentence Similarity Learning}
Sentence similarity learning is a fundamental and important NLP task which may be greatly enhanced by modeling the underlying semantic representations of compared sentences. In particular, a model should not be susceptible to variations of wording or syntax used to express the same idea. Moreover, a good model should also have the capacity to learn sentence similarity regardless of the length of the text and also needs to be efficient when applied to real-world applications~\cite{chen2017enhanced, enhancedrcnn}.

In the scene of insurance assessment, both efficiency and accuracy are of equal importance. Therefore we use the recent proposed Enhanced-RCNN model~\cite{enhancedrcnn} that achieves good performance on Chinese paraphrase identification datasets for learning sentence similarity. 

\subsection{Named Entity Recognition and Linking}
Named entity recognition is the NLP task of tagging entities in the text with the corresponding type and recent large-scale language model pretraining methods such as ELMo \cite{Peters:2018}, and BERT \cite{devlin2018bert} further enhanced the performance of NER, yielding state-of-the-art performances \cite{sutton2007dynamic, li2019unified}.

Entity linking is the task of recognizing and disambiguating named entities to a knowledge base \cite{hoffart2011robust, le2018improving}. EL can be split into two classes of approaches:
\begin{itemize}
\item \textbf{End-to-End:} processing a piece of text to extract the entities and then disambiguate these extracted entities to the correct entry in a given knowledge base.
\item \textbf{Disambiguation-Only:} directly takes gold standard named entities as input and only disambiguates them to the correct entry in a given knowledge base.
\end{itemize}

\subsection{Dialogue State Tracking}
Dialogue state tracking is an important component in task-oriented dialogue systems to identify users' goals and requests as a dialogue proceeds \cite{zhu2019sim}. The traditional DST system assumes that each slot's candidate values are within a limit number. However, this assumption does not apply to slots with an unlimited number of values in advance. It is more difficult for zero-shot domains like the scene of insurance assessment to predefine the slot values in advance \cite{ma2019end, ouyang2020dialogue}. 

Some recent researches are on converting fixed slot values into the substring of the dialogue context \cite{xu2018end, zhang2019find, gao2019dialog}. In this way, many researchers have proposed many neural networks to complete DST tasks in a reasonable way \cite{perez2017dialog, zhang2020recent}. Inspired by the recent progress, we also use the approach that treats the dialogue context as the source of slot values.

%% file: ethical.tex
Below we present the ethical considerations in terms of data authorization, privacy, and trust in real-world deployments.
\begin{itemize}
\item We have obtained explicit permissions from the claimant to use their personal private information data, including text and audio provided in the insurance assessment procedure for improving the system service.
\item The personal private information from the claimant is never used or stored for the commercial purpose.
\end{itemize}